%% file: lrec-coling2024.tex
\documentclass[10pt, a4paper]{article}
\usepackage{times}
\usepackage{latexsym}
\usepackage{graphicx}
\usepackage{xcolor}
\usepackage{array}
\usepackage{caption}
\usepackage{subcaption}
\usepackage{amsmath}
\usepackage{amsfonts}
\usepackage{amssymb}
\usepackage{bbm}
\usepackage{multirow}
\usepackage{float}
\usepackage{amsmath,amsfonts,amssymb}
\usepackage[]{lrec-coling2024} 

\usepackage{xcolor}
\usepackage{hyperref}
 \definecolor{darkblue}{rgb}{0, 0, 0.5}
  \hypersetup{colorlinks=true, citecolor=darkblue, linkcolor=darkblue, urlcolor=darkblue}

\usepackage{xstring}

\usepackage{color}

\title{LexAbSumm: Aspect-based Summarization of Legal Decisions}

\name{Santosh T.Y.S.S, Mahmoud Aly, Matthias Grabmair} 

\address{School of Computation, Information, and Technology; \\ Technical University of Munich, Germany
\\
\{santosh.tokala, mahmoud.aly, matthias.grabmair\}@tum.de\\}

\abstract{
Legal professionals frequently encounter long legal judgments that hold critical insights for their work. While recent advances have led to automated summarization solutions for legal documents, they typically provide generic summaries, which may not meet the diverse information needs of users. To address this gap, we introduce LexAbSumm, a novel dataset designed for aspect-based summarization of legal case decisions, sourced from the European Court of Human Rights jurisdiction. We evaluate several abstractive summarization models tailored for longer documents on LexAbSumm, revealing a challenge in conditioning these models to produce aspect-specific summaries. We release LexAbSum to facilitate research in aspect-based summarization for legal domain.
 \\ \newline  }

\begin{document}

\maketitleabstract

\section{introduction}
\input{text/Introduction}

\section{Dataset Construction}
\input{text/Dataset}

\section{Experiments}
\input{text/Experiments}

\section{Related Work}
\input{text/Relatedwork}

\section{Conclusion}
\input{text/Conclusion}

\section{Limitations}
\input{text/Limitations}

\section{Ethics Statement}
\input{text/Ethics}

\section{Bibliographical References}\label{sec:reference}

\bibliographystyle{lrec-coling2024-natbib}
\bibliography{lrec-coling2024}


\appendix
\input{text/appendix}

\end{document}

%% file: text/Introduction.tex

Legal professionals, including lawyers, judges, and researchers, regularly face the challenge of sifting through lengthy legal judgments that encompass multiple critical aspects for case law interpretation and judicial reasoning. Recent strides in NLP 
have addressed this challenge by offering automated legal case summarization systems (eg., \citealt{agarwal2022extractive, bhattacharya2021incorporating,bhattacharya2019comparative} \textit{inter alia}). These improvements owe much to the availability of datasets such as UK-Abs, IN-Abs, IN-Ext \cite{shukla2022legal}, BVA \cite{zhong2019automatic}, Rulingbr \cite{de2018rulingbr}, Multilexsumm \cite{shen2022multi} facilitating research in this field.


However, existing legal case summarization datasets and systems rely on a single, generic summary, which may not meet the diverse demands of users. Legal case documents are inherently multifaceted and often contain distinct aspects relevant to different users. This complexity is further amplified when documents involve multiple allegations or claimants. This one-size-fits-all approach risks omitting critical details and failing to provide specific information that individual users require \cite{angelidis2018summarizing,woodsend2012multiple}. Therefore, there is a need to develop legal case summarization systems capable of generating concise, aspect-specific summaries that cater to users' specific information needs more effectively.

Despite the undeniable demand for such systems, there currently exists no dataset designed explicitly for aspect-based legal case summarization. 
In this work, we present a novel dataset, named \textbf{LexAbSumm}\footnote{Our LexAbSumm dataset is available at \url{https://github.com/TUMLegalTech/LexAbSumm}}\enlargethispage{-\baselineskip} ,  tailored for evaluating single-document aspect-oriented abstractive summarization. 
LexAbSumm is sourced from the jurisprudence of the European Court of Human Rights (ECtHR), responsible for addressing complaints against states for alleged human rights violations under the European Convention of Human Rights. 
To create this dataset, we use theme-based fact sheets from the ECtHR's press service\footnote{\url{https://www.echr.coe.int/factsheets}}. From these sheets, we extract aspects from section titles 
and use the corresponding case details and brief summaries as aspect-based summaries of the case document. Our assessment of several abstractive summarization models, tailored for longer documents on LexAbSumm reveal that these models struggle to condition their summaries on the provided aspects, often producing generic summaries regardless of the aspect.

%% file: text/Dataset.tex

\textbf{Judgements Collection} 
We collect English judgments from the publicly accessible database of the ECtHR, HUDOC\footnote{\url{http://hudoc.echr.coe.int/}}, based on their metadata (Document Type: HEJUD). They follow a structured format, with sections like \emph{Procedure}, outlining the procedural steps; \emph{The Facts}, covering case background; \emph{The Law}, providing legal reasoning and \emph{Conclusion}, stating the Court's verdict on alleged violations. 
We use handcrafted rules and regular expressions to parse the documents  due to inconsistent HTML structure and section titles.
We exclude the \emph{Procedure} section as it focuses on procedural details. Throughout this work, we collectively refer to the combination of \emph{The Law} and \emph{Conclusion} as `the law section' since the conclusion primarily offers a final verdict based on the legal reasoning described in the Law section.
\newline

\begin{figure}
    \centering
    \includegraphics[width=0.4\textwidth]{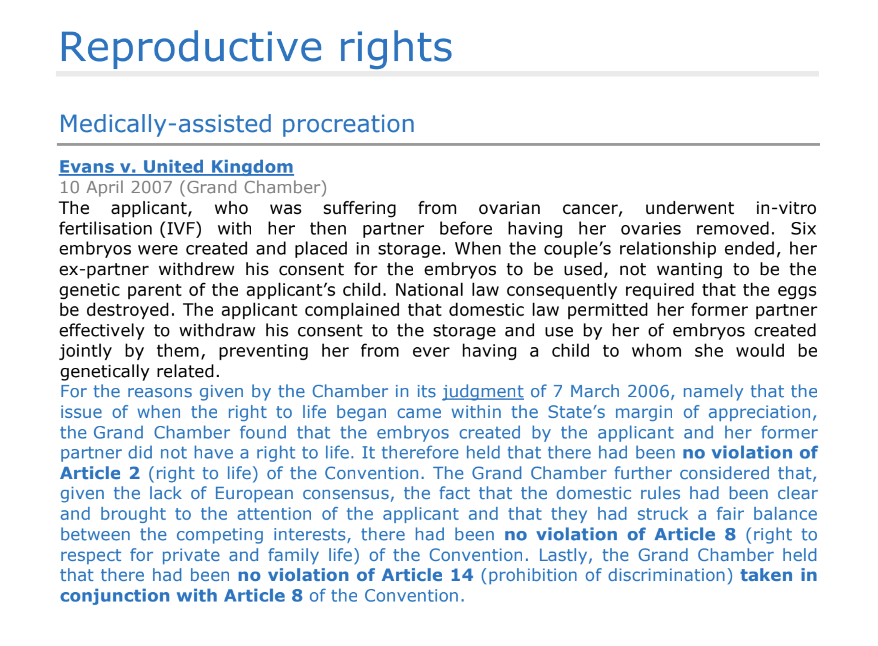}
    \caption{Example of an aspect-judgement-summary triplet from the fact sheet PDF file.}
    \label{summary-example}
\end{figure}

\noindent \textbf{Aspects and Summaries Collection} 
We curate our aspect-based summarization dataset using fact sheets available on the ECHR website. These sheets, prepared by the court's press service division, provide an analysis of case law developments across various thematic areas (e.g., Reproductive Rights \footnote{\url{https://www.echr.coe.int/documents/d/echr/\\FS\_Reproductive\_ENG}})
. In total, there are 73 fact sheets available as PDF documents, organized under 16 broad themes. Each sheet is further divided into sections (e.g., "Medically-assisted procreation" in the Reproductive Rights sheet), encompassing various aspects within a theme. Within each section, multiple cases and their summaries are provided. Each case summary includes two components: one summarizing the case's facts (in black font) and the other summarizing the case's legal aspects (in blue font). Fig. \ref{summary-example} illustrates an example of a case name with its summary and section title as presented in the PDF file. We employ various heuristics to extract section titles, cases, and their summaries from these PDF files. Subsequently, we manually filter out aspect-summary pairs in which either the summary is missing or the section title doesn't represent an aspect of the theme (e.g., "Pending applications"). Finally, we match case names in aspect-summary pairs back to our judgment collection and exclude pairs that couldn't be linked to our judgment collection, as some might refer to non-English documents not included in our collection.

\subsection{Dataset Splits \& Characteristics}

We end up with a total of 1053 aspect-judgement-summary triplets with 376 unique aspects. 
We randomly sample 7.5\% of these unique aspects (28) to create a test set of 91 triplets, allowing us to evaluate the models' generalization to new, unseen aspects. The remaining triplets are divided into training (810), validation (95), and test (57) sets. Notably, both the judgment and summary are available at a fine-grained level, segmented into facts and law sections. Based on this, we create three variants of aspect-based summarization task: using only the facts section, the law section and the whole (both the facts and the law) case as input, along with the aspect, to produce the corresponding summaries.

\begin{figure*}
  \centering
  \begin{subfigure}{0.3\textwidth}
    \includegraphics[width=\linewidth]{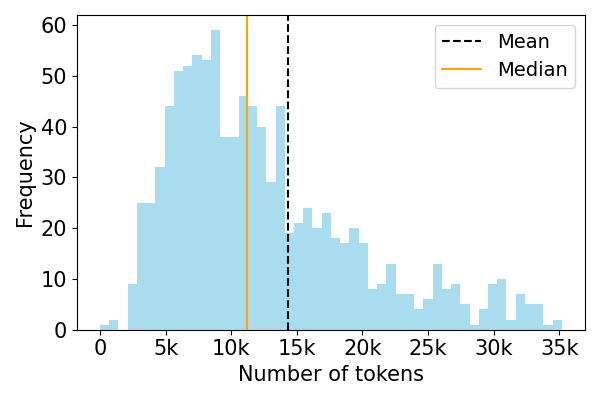}
    \caption{Distribution of number of tokens in the input of the whole split}
    \label{input-whole}
  \end{subfigure}
  \hfill
  \begin{subfigure}{0.3\textwidth}
    \includegraphics[width=\linewidth]{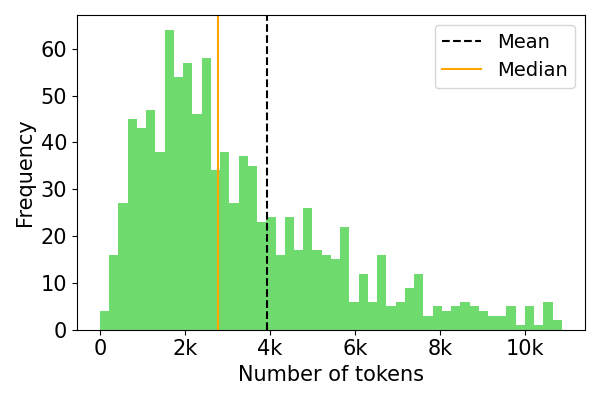}
    \caption{Distribution of number of tokens in the input of the facts split}
    \label{input-fact}
  \end{subfigure}
  \hfill
  \begin{subfigure}{0.3\textwidth}
    \includegraphics[width=\linewidth]{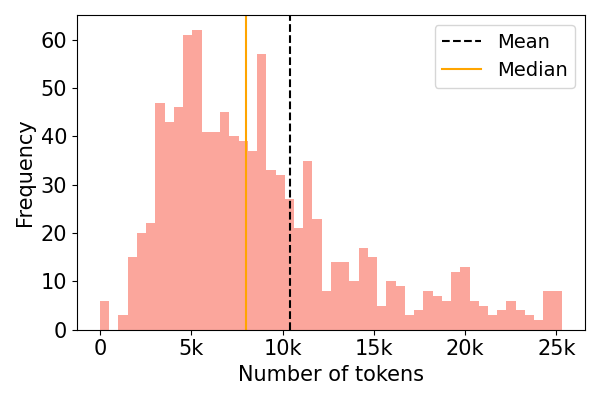}
    \caption{Distribution of number of tokens in the input of the law split}
    \label{input-law}
  \end{subfigure}

  \vspace{10pt} 

  \begin{subfigure}{0.3\textwidth}
    \includegraphics[width=\linewidth]{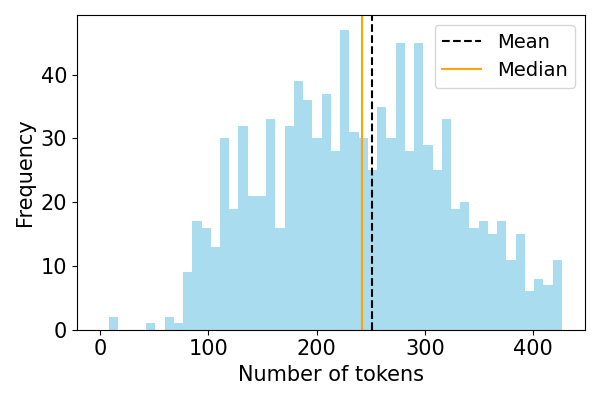}
    \caption{Distribution of number of tokens in the summary of the whole split}
    \label{output-whole}
  \end{subfigure}
  \hfill
  \begin{subfigure}{0.3\textwidth}
    \includegraphics[width=\linewidth]{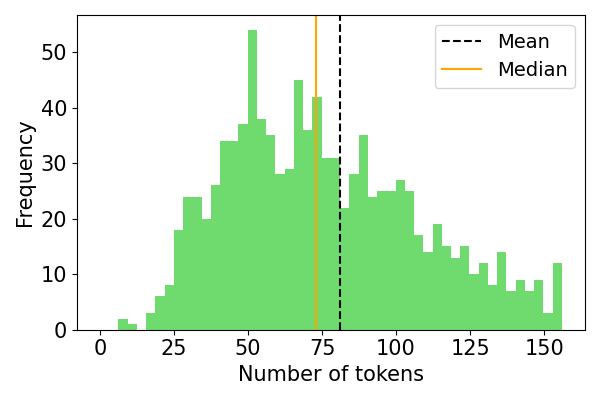}
    \caption{Distribution of number of tokens in the summary of the facts split}
    \label{output-fact}
  \end{subfigure}
  \hfill
  \begin{subfigure}{0.3\textwidth}
    \includegraphics[width=\linewidth]{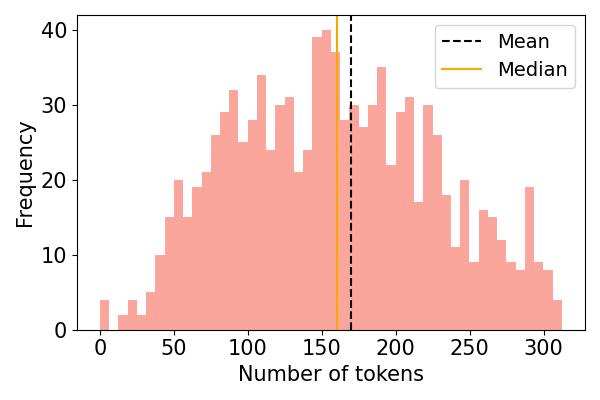}
    \caption{Distribution of number of tokens in the summary of the law split}
    \label{output-law}
  \end{subfigure}
    \caption{Distribution of input and output tokens in three LexAbSumm splits. }
  \label{token-inp-output}
\end{figure*}

\begin{figure}
    \centering
    \includegraphics[width=0.4\textwidth]{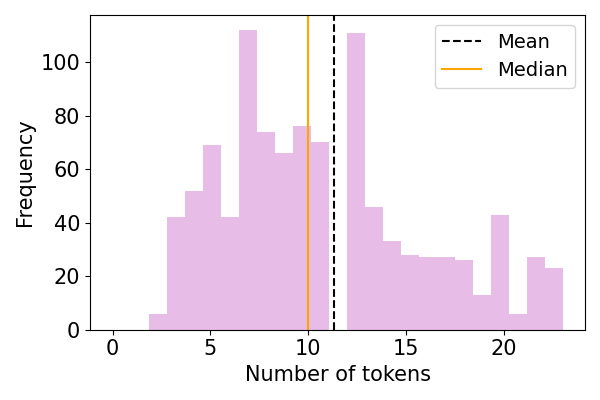}
    \caption{Distribution of tokens in the aspects of LexAbSumm.}
    \label{token-query-dist}
\end{figure}

\noindent \textbf{Data Analysis:} 
We report the following characteristics 
(i) Average number of tokens in the input and the summary. 
We also plot the token length distribution for the input and summary of three tasks in Fig. \ref{token-inp-output}. 
(ii) Compression Ratio indicates the token ratio between the input and the summary. (iii) Coverage-n \cite{grusky2018newsroom} indicates the \% of n-grams in the summary that are part of an extractive fragment within the input and quantifies the extent to which a summary is derivative of an input text. (iv) Density-n \cite{grusky2018newsroom} quantifies how well the n-gram sequence of a summary can be described as a series of extractions and is derived from the average length of the extractive fragment to which each n-gram in the summary belongs. (v) Copy Length \cite{chen2020cdevalsumm} denotes the average length of segments in summary copied from the input. (vii) Novelty-n \cite{see2017get} denotes the ratio of new n-grams present in the summary that are not in the input. (viii) We plot the distribution of aspect length tokens in Fig. \ref{token-query-dist}.

\input{text/tab-data-char}

Table \ref{datachar-tab} shows that the law sections are longer than the facts sections in both the input and output, even with higher compression ratios, requiring more precise capture of crucial points from the input with respect to aspect presenting a greater challenge.  
We notice that the data exhibits higher coverage and density at the unigram level, suggesting higher token-level extractiveness. However, this does not necessarily translate to overall extractiveness, as it tends to rise with longer input documents  that contain various summary tokens dispersed throughout.
At the bigram level, both coverage and density drop,  indicating token dispersion. This observation is further supported by the copy length metric. These metrics are notably higher for the law section compared to the facts, due to the specific vocabulary used in legal reasoning, tests and principles which leave little room for paraphrasing, while the progression of events in the facts are more adaptable to paraphrasing. This is also evident in higher novelty for facts compared to law. While unigram novelty may appear smaller, it increases with n-grams, indicating that the summaries are highly abstractive, requiring synthesis of information from multiple segments across the input. Average number of tokens in aspect is 10.2. 


%% file: text/tab-data-char.tex
\begin{table}[]
\scalebox{0.78}{
\centering
\begin{tabular}{|l|c|c|c|}
\hline
& \textbf{Whole} & \textbf{Facts} & \textbf{Law} \\  \hline
\textbf{Input Length}      & 14357.14       & 3929.77        & 10427.38     \\ 
\textbf{Summ. Length}      & 251.1          & 81.19          & 169.91       \\ 
\textbf{Comp. Ratio}       & 66.25          & 59.75          & 85.02        \\ 
\textbf{Coverage (1/2-gram)} & 0.95/0.73           & 0.87/0.46           & 0.96/0.74         \\ 
\textbf{Density (1/2-gram)}  & 6.32/5.18         & 2.56/1.49           & 7.08/5.94         \\ 
\textbf{Copy Length}       & 1.98           & 1.53           & 2.01         \\ 
\textbf{Novelty (1/2-gram)}  & 0.07/0.49       & 0.26/0.75         & 0.08/0.45      \\ 
\textbf{Novelty (3-gram)}  & 0.7            & 0.9            & 0.66         \\ \hline
\end{tabular}}
\caption{Characteristics of LexAbSumm.}
\label{datachar-tab}
\end{table}

%% file: text/Experiments.tex
\subsection{Models \& Metrics}
We evaluate the following abstractive models designed to handle longer inputs on our three task variants. These are structured as sequence-to-sequence tasks, where the aspect is prepended to the document with a delimiter as the sequence input and the summary as output.


\noindent \textbf{LED} \cite{beltagy2020longformer} is a longformer variant equipped with both encoder and decoder. In the encoder, it uses efficient local+global attention pattern instead of full self-attention, while the decoder utilizes full quadratic attention. It is initialized from the pre-trained BART model \cite{lewis2020bart}. We use the LED-base version, which can handle input lengths of up to 16,384 tokens.


\noindent \textbf{PRIMERA} \cite{xiao2022primera} is initialized with the LED model and pre-trained with a summarization specific entity-based sentence masking objective and can handle upto 4096 tokens.

\noindent \textbf{LongT5} \cite{guo2022longt5} 
uses 
local+global attention from ETC \cite{ainslie2020etc} and adopt summarization specific pre-training from PEGASUS into the T5 model \cite{raffel2020exploring} for longer sequences. We use the LongT5-base version which can handle upto 16384 tokens.

\noindent \textbf{SLED} \cite{ivgi2023efficient} processes long sequences by partitioning the input into overlapping chunks and encode each chunk with a short-range pre-trained models encoder. It relies on decoder to fuse information across chunks attending to all input tokens, similar to fusion-in-decoder \cite{izacard2021leveraging}. SLED can be applied on top of any short-range models and thus we derive SLED-BART-base by applying on BART-base. 

\noindent \textbf{Unlimiformer} \cite{bertsch2023unlimiformer} adopts a strategy similar to SLED, but rather than attending to all input tokens, it focuses exclusively on the top-k tokens retrieved from a k-nearest-neighbor index constructed over the hidden states of all input tokens at every decoder layer. 
This design helps to handle unbounded length during testing, in contrast to SLED, which is limited by memory when attending to all input tokens in the decoder. We create the Unlimiformer-BART-base model.

\noindent For both SLED and Unlimiformer, we append the aspect to every chunk of the input. 

\noindent \textbf{Metrics:} We calculate ROUGE-\{1,2,L\} \cite{lin2004rouge} and BERTScore (BS) \cite{zhang2019bertscore} to compute the lexical and semantic overlap between the generated and the golden summaries for each instance and report the averaged values across all the 148 test (seen + unseen) instances.

\subsection{Implementation Details}
We implement our code using Huggingface Transformers library \cite{wolf2020transformers} 
. We sweep over learning rates $\{1e-5, 3e-5, 5e-5, 1e-4, 3e-4\}$ and select the best model based on the R-L score on the validation set. The model is trained for 10 epochs with Adam optimizer \cite{kingma2014adam}. The inputs and outputs are truncated to a maximum of 16384/512 tokens for all models except PRIMERA, which is 4096/512 tokens. SLED and Unlimiformer have a chunk size of 256, and a chunk overlap of 0.5. In Unlimiformer, 16k tokens are used at training time, while unlimited tokens at evaluation.

\input{text/tab-results}

\subsection{Results}

We use three extractive baselines: 
\textbf{Random} selects input sentences randomly as the summary. \textbf{Similarity} chooses sentences similar to the query using cosine similarity with BERT embeddings. \textbf{Extractive Oracle} is a greedy algorithm that iteratively selects sentences which maximize ROUGE-2 with the reference summary, usually regarded as the the \textit{upper bound} of the extractive method \cite{nallapati2017summarunner}. These baselines use the same number of sentences as the reference summary.

Table \ref{results-tab} indicates that abstractive models outperform the Random and Similarity baselines. Among the abstractive models, LED outperforms PRIMERA, due to its longer input capacity (16k vs. 4k), evident in R-L scores for the law and whole variants with longer inputs. 
LongT5 surpasses them all, benefiting from its end-to-end pre-training for lengthy sequences. 
SLED and Unlimiformer outperform all, suggesting that adapting short-range pre-trained models into those frameworks boost performance.  Unlimiformer is comparable or better to SLED 
demonstrating that attending to the top-k input keys can be an accurate approximation of full attention.
The Extractive Oracle, which leverages target summaries, serves as a strong baseline, emphasizing the potential of an extract-then-summarize paradigm in future.

\input{text/tab-analysis}

\subsection{Analysis}
\noindent \textbf{Generalizability to new aspects}
We report the models' performance separately for triplets covering seen (57 instances) and unseen aspects (91 instances) during training using the whole split in Table \ref{analysis-tab}. 
We notice a decline in all models when handling unseen aspects. 
Future work should focus on enhancing model robustness to new aspects, especially in the legal domain, where the evolution of case law and legal norms requires adaptability to novel aspects.

\noindent \textbf{Sensitivity to aspects}
We assess the model's aspect sensitivity by examining triplets in the test set where the judgment document remains the same, but the aspects differ. We calculate the BLEU score \cite{papineni2002bleu} between every pair of summaries (42 pairs) with the same case and report the average in Table \ref{analysis-tab}. 
Lower BLEU scores indicate the model's ability to generate distinct summaries for different aspects rather than a generic one, with a lower bound set by the Oracle score.
We observe that long-range models like LED and LongT5 tend to produce general summaries with less aspect conditioning due to the longer positional distance between the aspect and the input. 
In contrast, SLED mitigates this effect by concatenating the aspect as a prefix to every chunk. However, this concatenation effect is subdued by top-k attention in Unlimiformer. Although PRIMERA exhibits better aspect sensitivity, it lags in ROUGE and BS, indicating a room for improvement in effectively conditioning on the aspect, especially for longer documents.


%% file: text/tab-results.tex
\begin{table*}[ht]
\scalebox{0.85}{
\centering
\begin{tabular}{|l||l|l|l|l||l|l|l|l||c|c|c|c|}
\hline
 & \multicolumn{4}{c||}{\textbf{Whole}}                                                                                                          & \multicolumn{4}{c||}{\textbf{Facts}}                                                                                                          & \multicolumn{4}{c|}{\textbf{Law}}                                                                                                            \\ \hline
\textbf{Models}        & {\textbf{R-1}} & {\textbf{R-2}} & {\textbf{R-L}} & {\textbf{BS}} & {\textbf{R-1}} & {\textbf{R-2}} & {\textbf{R-L}} & {\textbf{BS}} & {\textbf{R-1}} & {\textbf{R-2}} & {\textbf{R-L}} & {\textbf{BS}} \\ \hline
\textbf{Random}      & {41.24}        & {10.94}        & {19.29} & {82.77} & {25.85}        & {4.99}         & {15.06} & {78.98} & {36.13}        & {9.38}         & {18.98} & {82.63} \\ 
\textbf{Similarity}  &  44.24 & 14.88 &	22.11& 83.58 &	31.23 &	7.6	 & 18.22 & 	83.89 &	41.32 &	14.07	&22.45	&84.14                      \\ \hline
\color[HTML]{9B9B9B} \textbf{Ext. Oracle} &  \color[HTML]{9B9B9B}68.27	& \color[HTML]{9B9B9B}46.46 &	\color[HTML]{9B9B9B}34.11 &	\color[HTML]{9B9B9B}89.1 &	\color[HTML]{9B9B9B}46.22 &	\color[HTML]{9B9B9B}22.95 &	\color[HTML]{9B9B9B}26.67 &	\color[HTML]{9B9B9B}86.97 & 	\color[HTML]{9B9B9B}68.48 &	\color[HTML]{9B9B9B}48.59	& \color[HTML]{9B9B9B}39.57 & 	\color[HTML]{9B9B9B}90.04 \\ \hline
\textbf{LED}         & {49.56}        & {25.53}        & {30.28} & {87.28} & {40.43}        & {15.38}        & {26.36} & {88.09} & {47.08}        & {26.82}        & {33.49} & {88.41} \\ 
\textbf{PRIMERA}     & {47.90}         & {22.03}        & {27.89} & {86.94} & {38.71}        & {14.29}        & {25.28} & {87.81} & {46.09}        & {21.69}        & {29.04} & {87.25} \\ 
\textbf{Long-T5}     & {50.91}        & {26.51}        & {31.27} & {87.72} & {40.68}        & {15.74}        & {26.83} & {88.37} & {47.64}        & {27.30}         & {33.86} & {88.85} \\ 
\textbf{SLED-BART}   & 52.41                            & 28.28                            & 32.97                     & 88.33                     &    40.61 & 15.91	& 26.26 &	87.99  &    49.24 & 29.73 & 35.34 & 88.97          \\ 
\textbf{Unlim.-BART}   & 51.53 & 27.77 &	32.28 & 88.21   & 40.67 & 	16.04 &	26.45	& 88.27  & 50.62 & 30.28 &	35.55 & 88.86      \\ 
\hline
\end{tabular}}
\caption{Results on three LexAbSumm datasets. Extractive Oracle (Grey) leverages target summaries.}
\label{results-tab}
\end{table*}

%% file: text/tab-analysis.tex
\begin{table*}[]
\centering
\scalebox{0.85}{
\begin{tabular}{|l||l|l|l|l||l|l|l|l|l|}
\hline
                     & \multicolumn{4}{c|}{\textbf{Seen}}                                                                                      & \multicolumn{4}{c|}{\textbf{Unseen}}  &   \multirow{2}{*}{\textbf{BLEU}}                                                                                  \\ \cline{1-9}
\textbf{Models}      & \multicolumn{1}{c|}{\textbf{R-1}} & \multicolumn{1}{c|}{\textbf{R-2}} & {\textbf{R-L}} & \textbf{BS} &  \multicolumn{1}{c|}{\textbf{R-1}} & \multicolumn{1}{c|}{\textbf{R-2}} & {\textbf{R-L}} & \textbf{BS} & \\ \hline
\textbf{Oracle}   & \multicolumn{1}{l|}{}             &         & & & &                         & \multicolumn{1}{l|}{}             &                                  & 21.06                                                                         \\ \hline
\textbf{LED}         & {49.04}             & {26.68}             & {31.3}             &     {88.01}      & {48.81}  & {24.82}             & {29.64}             & {87.43}             &       {46.92}      \\ 
\textbf{PRIMERA}     & {50.16}             & {22.74}             & {28.16}             &       {86.88}      & {48.18}   & {21.59}           & {27.71}             & {86.98}             &      {31.70}       \\ 
\textbf{Long-T5}     & {51.10}             & {27.95}             & {32.68}             &      {88.42}       & {50.79}   & {25.61}           & {30.39}             & {87.84}             &        {51.07}     \\ 
\textbf{SLED-BART}   & {54.17}             & {29.80}             & {34.14}             &      {88.84}       & {51.31}   & {27.32}           & {32.24}             & {88.27}             &      {42.12}       \\ 
\textbf{Unlim.-BART} & {52.49}             & {29.72}             & {34.55}             &      {88.55}       & {50.93}   & {26.55}           & {30.86}             & {87.99}             &      {54.41}       \\ 
\hline
\end{tabular}}
\caption{Anlysis on the Whole variant of LexAbSumm. `Seen’ and ‘Unseen’ column depict new aspect generalizability, ‘BLEU’ column depict aspect sensitivity.}
\label{analysis-tab}
\end{table*}

%% file: text/Relatedwork.tex

\noindent \textbf{ECtHR as data resource}
This has been utilized in various NLP tasks including judgement prediction \cite{aletras2016predicting,chalkidis2019neural,chalkidis2021paragraph,santosh2022deconfounding,tyss2023zero,tyss2023leveraging,xu2023dissonance}, argument mining \cite{habernal2023mining,poudyal2019using,poudyal2020echr}, vulnerability detection \cite{xu2023vechr}, event extraction \cite{filtz2020events,navas2022whenthefact}.
While these tasks center around the judgment corpus, we utilize the press release fact sheets from the ECtHR to develop our aspect-based summarization dataset, LexAbSumm.

\noindent \textbf{Legal Case Summarization}
There exist datasets from various jurisdictions such as HOLJ  \cite{grover2004holj} and UK-Abs from UK, IN-Abs and IN-Ext from India \cite{shukla2022legal}, BVA from US \cite{zhong2019automatic}, CanLII from Canada \cite{zhong2022computing}, Rulingbr  \cite{de2018rulingbr} from Brazil, 
Multilexsumm \cite{shen2022multi} from civil rights cases of US. Traditional approaches have leaned on extractive algorithms, known for their faithfulness \cite{zhong2019automatic,bhattacharya2019comparative}, while the recent ones explored abstractive methods \cite{shukla2022legal,shen2022multi,elaraby2022arglegalsumm}. These datasets provide single, generic summary, which may not meet differentiated user needs. In contrast, our work pioneers aspect-based legal case summarization. 

\noindent \textbf{Aspect-based Summarization}
Initial works were based on 
aspects in opinions and reviews \cite{woodsend2012multiple,angelidis2018summarizing,wu2016aspect} which got extended to 
news \cite{ahuja2022aspectnews,bahrainian2022newts,frermann2019inducing}, scientific documents \cite{meng2021bringing} and wikipedia \cite{hayashi2021wikiasp,yang2022oasum}. Closely related area is query-based summarization \cite{kulkarni2020aquamuse,zhong2021qmsum,wang2022squality} focusing on specific fact or knowledge based questions 
instead of high-level aspects which mimic user intents. 
We introduce LexAbSumm, an aspect-based summarization data for the legal domain. 

%% file: text/Conclusion.tex



We introduce LexAbSumm, the first aspect-based summarization dataset for legal judgments, sourced from ECtHR fact sheets. Unlike traditional summarization, LexAbSumm targets differentiated summaries based on user needs (aspects). We evaluate abstractive models tailored for longer inputs, highlighting their limitations in aspect conditioning, indicating room for improvement. We hope our data and models can foster research in this area. In future, we intend to develop a model that not only conditions on the given aspect but also
discourages the generation of non-aspect-related words during decoding. 

%% file: text/Limitations.tex
While our work has introduced a novel aspect-based summarization dataset for legal judgments, it is essential to acknowledge its limitations:

\noindent \textbf{Language and Jurisdiction Specificity:} LexAbSumm is primarily derived from legal judgments within the jurisdiction of the European Court of Human Rights (ECtHR). As such, the language, legal principles, and specific nuances in these judgments may not be directly transferable to other jurisdictions, limiting its generalizability. Further evaluation need to be carried out to understand the cross domain generalization abilities. 

\noindent \textbf{Evaluation Metrics:} The paper primarily uses standard NLP evaluation metrics like ROUGE and BLEU. While these metrics provide useful insights, they may not fully capture the nuanced quality of legal summaries, and developing domain-specific evaluation metrics could be beneficial. 

\noindent \textbf{Aspects Coverage:} The dataset's aspects are constructed based on fact sheet section titles. This approach may not comprehensively cover all the potential aspects relevant to legal professionals, as it relies on the available thematic organization within the fact sheets. Future work could explore methods for incorporating a wider range of aspects, reflecting diverse legal cases and user needs.

%% file: text/Ethics.tex
We anticipate no ethical issues in redistributing our dataset as a publicly accessible resource. This is because the legal judgment data and summaries extracted from the fact sheets are already available through HUDOC and the ECtHR website, and our utilization aligns with the ECtHR data policy. While it's worth noting that these judgments contain real names of individuals, our approach does not introduce any harmful use beyond their existing public accessibility. Our primary goal is to encourage further research in the realm of Legal NLP, which currently suffers from a scarcity of data. Our aim is to create technology that assists legal professionals. It's important to acknowledge that our use of pre-trained language models may inherit biases from their training data. We emphasize the need to thoroughly investigate and address any biases that may emerge, ensuring the fairness and ethical usage of the systems we develop.


%% file: text/appendix.tex

\section{Appendix}
\subsection{Models Performance Analysis}
Tables \ref{tab-analysis-facts} and \ref{tab-analysis-law} provided the detailed results on generlizability to new aspects and sensitivity to aspects for the facts and law variants, across all the models respectively.

\input{text/tab-appendix}

%% file: text/tab-appendix.tex
\begin{table*}[]
\centering
\begin{tabular}{|l||l|l|l|l||l|l|l|l||l|}
\hline
                     & \multicolumn{4}{c||}{\textbf{Seen}}                                                                                      & \multicolumn{4}{c||}{\textbf{Unseen}}  &   \multirow{2}{*}{\textbf{BLEU}}                                                                                  \\ \cline{1-9}
\textbf{Models}      & \multicolumn{1}{c|}{\textbf{R-1}} & \multicolumn{1}{c|}{\textbf{R-2}} & {\textbf{R-L}} & \textbf{BS} & \multicolumn{1}{c|}{\textbf{R-1}} & \multicolumn{1}{c|}{\textbf{R-2}} & {\textbf{R-L}} & \textbf{BS} & \\ \hline
\textbf{LED}         & {40.73}             & {15.78}             & {26.87}             &      {88.04}       & {40.57}  & {15.13}            & {26.69}             & {88.12}             &      {39.4}       \\ 
\textbf{PRIMERA}     & {38.82}             & {15.64}             & {25.83}             &     {87.69}        & {38.64}  & {13.38}            & {24.94}             & {87.89}             &     {32.78}        \\ 
\textbf{Long-T5}     & {41.12}             & {15.91}             & {26.78}             &       {88.52}      & {40.91}  & {15.62}            & {26.63}             & {88.28}             &       {61.42}      \\ 
\textbf{SLED-BART}   & {41.73}             & {19.38}             & {29.81}             &     {88.61}        & {36.67}  & {12.6}            & {24.04}             & {87.6}             &     {57.1}        \\ 
\textbf{Unlim.-BART} & {41.94}             & {19.67}             & {29.46}             &     {88.66}        & {38.25}  & {13.77}            & {24.57}             & {88.02}             &       {34.94}      \\ \hline
\end{tabular}
\caption{Anlysis on the facts varirant of LexAbSumm. `Seen’ and ‘Unseen’ column depict new aspect generalizability, ‘BLEU’ column depict aspect sensitivity.}
\label{tab-analysis-facts}
\end{table*}

\begin{table*}[]
\centering
\begin{tabular}{|l||l|l|l|l||l|l|l|l||l|}
\hline
                     & \multicolumn{4}{c||}{\textbf{Seen}}                                                                                      & \multicolumn{4}{c||}{\textbf{Unseen}}  &   \multirow{2}{*}{\textbf{BLEU}}                                                                                  \\ \cline{1-9}
\textbf{Models}      & \multicolumn{1}{c|}{\textbf{R-1}} & \multicolumn{1}{c|}{\textbf{R-2}} & {\textbf{R-L}} & \textbf{BS} & \multicolumn{1}{c|}{\textbf{R-1}} & \multicolumn{1}{c|}{\textbf{R-2}} & {\textbf{R-L}} & \textbf{BS} &  \\ \hline
\textbf{LED}         & {47.52}             & {28.91}             & {35.36}             &      {88.49}       & {46.58}  & {25.51}             & {32.31}             & {88.24}             &       {53.01}      \\ 
\textbf{PRIMERA}     & {45.28}             & {21.65}             & {28.71}             &      {87.09}       & {45.98}  & {21.72}             & {28.71}             & {87.35}             &        {27.01}     \\ 
\textbf{Long-T5}     & {47.8}             & {29.53}             & {36.09}             &     {88.79}        & {46.91}  & {25.91}             & {32.53}             & {88.53}             &      {86.56}       \\ 
\textbf{SLED-BART}   & {50.66}             & {32.25}             & {37.51}             &       {89.32}      & {48.35}  & {28.15}            & {33.98}             & {88.76}             &      {41.35}       \\ 
\textbf{Unlim.-BART} & {51.44}             & {31.86}             & {37.52}             &      {89.09}       & {50.1}    & {29.3}          & {34.31}             & {88.73}             &      {61.44}       \\ \hline

\end{tabular}
\caption{Anlysis on the Law varirant of LexAbSumm. `Seen’ and ‘Unseen’ column depict new aspect generalizability, ‘BLEU’ column depict aspect sensitivity.}
\label{tab-analysis-law}
\end{table*}